\documentclass[runningheads]{llncs}

\usepackage{eccv}

\usepackage{eccvabbrv}

\usepackage{graphicx}
\usepackage{booktabs}
\usepackage{multirow}

\usepackage[accsupp]{axessibility}  

\usepackage{xcolor}
\usepackage{pifont}

\usepackage{tikz}

\definecolor{algogreen}{HTML}{499B4A}
\definecolor{algored}{HTML}{D12A2F}

\usepackage{hyperref}

\usepackage{orcidlink}

\begin{document}

\title{A Framework for Egocentric and Exocentric Procedural Understanding via Temporal Segmentation and Semantic Abstraction.} 

\titlerunning{Egocentric and Exocentric Procedural Understanding}

\author{
Vivek Chavan\inst{1,2}\orcidlink{0000-0001-9350-5259} \thanks{Project Lead. Correspondence: \texttt{\href{mailto:contact@vivekchavan.com}{contact@vivekchavan.com}}} \and
Jörg Krüger\inst{1,2}\orcidlink{0000-0001-5138-0793} 
}

\authorrunning{V. Chavan et al.}

\institute{
Fraunhofer IPK, Berlin, Germany\\
\and
Technische Universität Berlin, Berlin, Germany
}

\maketitle

\begin{abstract}
Long-horizon ego/exo data contains rich procedural evidence, but are redundant, noisy, and costly to process or retain. We propose a compact framework that converts continuous multimodal workplace video into a structured Procedural State Memory, implemented as a Work Environment Model (WEM). Inspired by event segmentation theory, we detect boundaries using changes in visual context, location, motion, narration, gaze/object interaction, and optional exocentric workspace evidence, rather than fixed windows or visual novelty alone. Each segment is abstracted into an evidence-linked event card containing actor, interval, location, action, objects/tools, pre/post state, confidence, and provenance. These event cards incrementally update the WEM, enabling compact, auditable documentation and retrieval under on-premise privacy constraints. We instantiate the design with frozen DINOv2 and VJEPA-2 encoders and a local language model, and outline evaluation criteria for segmentation quality, memory compression, retrieval fidelity, and long-horizon QA.
\keywords{Egocentric vision \and Ego--exo procedural understanding \and Event segmentation \and Structured memory \and Industrial video understanding}
\end{abstract}

\section{Introduction}

Egocentric perception captures activity from the viewpoint of the acting user, making it well suited for understanding hands, tools, gaze, narration, and task context \cite{engel2023projectarianewtool, ego4d, plizzari2024outlook}. Exocentric views provide complementary information about workspace layout, collaborators, and object end states that may be occluded or outside the wearable field of view \cite{egoexo4d, charadesego}. Together, ego--exo streams offer a rich basis for long-horizon procedural understanding in real-world work.

However, continuous procedural video cannot be handled simply by increasing context length or captioning fixed windows. Most frames are redundant, task relevance is sparse, and important information often lies in state changes: where an object was placed, whether a tool was used, which step was completed, or which condition changed. This motivates the view of procedural understanding as \emph{continual procedural learning}: constructing, updating, and querying task-relevant state from continuous multimodal experience.

We build on event segmentation theory, which suggests that humans parse continuous experience into events when the current event model no longer explains incoming observations \cite{zacks2007eventsegmentation}. In our setting, this means that event boundaries should reflect changes in procedural state, not only changes in visual appearance. We therefore combine cues from scene features, location, motion, narration, gaze/object interaction, and exocentric workspace evidence to propose meaningful temporal units.

Industrial environments are a useful testbed because they contain repeated procedures, shared tools, known locations, and safety constraints, but also require privacy, data minimisation, and on-premise processing \cite{Chavan2025IndEgo, chavan2024applicationegocentriccomputervision}. We focus on the memory representation needed by such systems. Our framework converts raw ego/exo recordings into structured event records and updates a Work Environment Model containing locations, objects, tools, procedures, evidence, uncertainty, and dynamic object/task state.

\section{Related Work}
\label{sec:related_work}

\noindent\textbf{Egocentric and ego--exo procedural understanding.}
Egocentric video captures the actor's hands, gaze, narration, tools, and immediate task context, making it central to procedural video understanding \cite{wearcomp, 2012_ego, ego_stanford, ego4d, epic_kitchens}. Recent datasets and benchmarks extend this setting to long-horizon activity, episodic memory, hand-object interaction, and paired ego--exo observations \cite{ego4d, egoexo4d, hot3d}. However, first-person video remains difficult due to motion blur, occlusion, rapid viewpoint change, and sparse task relevance \cite{engel2023projectarianewtool, ego_stanford, recogniseegocentric}. Our work focuses on the industrial case, where repeated procedures and shared tools provide structure, but privacy and data efficiency constrain how much raw video can be retained \cite{Chavan2025IndEgo, chavan2024applicationegocentriccomputervision}.

\noindent\textbf{Temporal segmentation and summarisation.}
Prior egocentric work studies video summarisation, action recognition, gaze, and hand-object interaction \cite{egocentric_summarization, summarisation_predicting, gaze_egocentric, hands_egocentric}. Many practical pipelines still rely on fixed temporal windows, generic video captions, or single-cue scene changes. In contrast, event segmentation theory suggests that meaningful boundaries occur when the current event model no longer explains incoming observations \cite{zacks2007eventsegmentation}. We adopt this view for procedural video: boundaries should reflect changes in task state, combining visual novelty, place, motion, gaze/object interaction, narration, and ego--exo evidence.

\noindent\textbf{Multimodal assistants and structured memory.}
Multimodal assistants and VLMs can process images, audio, text, and video \cite{openai2024gpt4technicalreport, openai2024gpt4o, liu2023llava}, and egocentric assistants have been explored for spatial grounding, intent inference, and procedural guidance \cite{learn_spatial, jia2022egotaskqa, HoloAssist2023, yang2025egolifeegocentriclifeassistant}. Yet industrial use requires on-premise processing, auditable outputs, and selective retention of sensitive data \cite{vyatkin2013software}. Rather than treating a large model as the sole memory, we store compact event records in a Work Environment Model linking locations, tools, actions, procedures, timestamps, confidence, and evidence. This provides a queryable procedural memory while reducing dependence on raw-video retention.

\section{Methods}
\label{sec:methods}

\begin{figure*}[t]
    \centering
    \includegraphics[width=\textwidth]{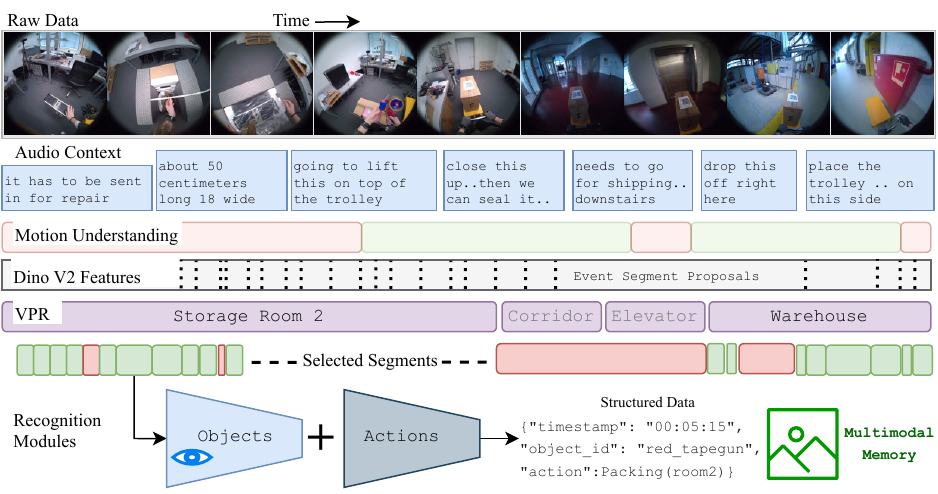}
    \caption{\textbf{Framework overview.} Ego--exo recordings are processed offline with frozen encoders to extract visual, motion, gaze/object, narration, location, and exocentric cues. A multi-cue boundary score segments the stream into procedural events. Each event is converted into an evidence-linked event card and inserted into a Work Environment Model (WEM), which supports compact documentation, retrieval, and QA.}
    \label{fig:method}
\end{figure*}

We develop the framework using the IndEgo dataset \cite{Chavan2025IndEgo} and Project Aria tools \cite{engel2023projectarianewtool}. We target \emph{offline, on-premise} processing: raw recordings are transferred to a local workstation, processed in batches, and compressed into structured memory. The goal is not a full real-time assistant, but the representational core for long-horizon procedural understanding: segmentation, abstraction, and memory update.

\noindent\textbf{Bootstrapping the Work Environment Model.}
Before deployment, subject-matter experts provide lightweight workplace priors: location exemplars for visual place recognition, object/tool exemplars and aliases, action prototypes, and procedure templates. These initialise a Work Environment Model (WEM), represented as a graph
$G=(V,E)$ with nodes for \emph{locations, objects/tools, actions, procedures, users, and events}. Edges encode \emph{observed\_at, manipulates, performed\_by, precedes, part\_of\_procedure}, and \emph{has\_state}. Each node/edge stores timestamps, confidence, source model, and evidence IDs. We use frozen foundation encoders and local indices/prototypes rather than full model fine-tuning \cite{dinov2, vjepa2, llama}.

\noindent\textbf{Multi-cue temporal segmentation.}
Given an ego--exo stream, we extract sparse per-time-step cues
$\mathcal{C}=\{g,o,m,a,l,x\}$ for global visual appearance, object/gaze interaction, motion, audio/narration, location, and optional exocentric context. DINOv2 features provide global visual and VPR cues \cite{dinov2, vpr}; pose/IMU gives motion transitions; gaze fixations restrict object evidence to attended regions; transcripts provide topic or instruction shifts; exocentric frames provide stable workspace evidence when available. For cue $c$, novelty is computed against an exponential moving prototype:
\begin{equation}
n_t^c = 1-\mathrm{sim}(z_t^c,\mu_{t-1}^c), \qquad
N_t = \sum_{c\in\mathcal{C}_t} \bar{w}_c\,\rho(n_t^c),
\label{eq:novelty}
\end{equation}
where $z_t^c$ is the cue feature, $\mu_{t-1}^c$ its running prototype, $\rho$ is robust normalisation, and $\bar{w}_c$ renormalises weights over available modalities. Boundaries are selected where
\begin{equation}
t \in \mathcal{B} \quad \text{if} \quad
N_t > \mathrm{median}(N) + k\,\mathrm{MAD}(N),
\label{eq:boundary}
\end{equation}
followed by non-maximum suppression and a minimum segment duration constraint. This yields variable-length procedural segments rather than fixed windows.

\noindent\textbf{Semantic abstraction.}
Each segment $S_i=[t_i,t_{i+1})$ is passed to specialised recognition modules. DINOv2 \cite{dinov2} supports location and object/tool grounding; VJEPA-2 \cite{vjepa2} encodes temporal dynamics for action recognition; narration and gaze provide cheap disambiguating cues. The result is an event card
\begin{equation}
E_i=\langle u,S_i,l_i,a_i,O_i,s_i^{-},s_i^{+},\gamma_i,p_i\rangle ,
\end{equation}
containing user, interval, location, action, objects/tools, pre/post state, confidence, and provenance. A local Llama model normalises labels to the WEM vocabulary, resolves aliases, and produces short textual summaries, but every generated statement remains linked to timestamps and evidence frames \cite{llama}.

\noindent\textbf{Memory update and retrieval.}
Event cards incrementally update the WEM via $G_{i+1}=U(G_i,E_i)$: new event nodes are inserted, object/tool states are updated, and temporal/procedural edges are added. The WEM maintains private per-user memory and a reviewed shared layer for workplace-level facts such as tool locations, procedure steps, and common object states. Queries retrieve relevant subgraphs by time, location, object, action, or procedure; the local language model then generates answers or summaries from retrieved WEM entries rather than from the full raw video. This design makes procedural memory compact, auditable, and compatible with industrial privacy constraints.

\section{Conclusion and Outlook}
\label{sec:conclusion}

The proposed pipeline identifies multi-cue event boundaries and converts each segment into an evidence-linked procedural state record. These update a Work Environment Model that stores actions, objects, locations, state changes, confidence, timestamps, and provenance, providing a compact and auditable memory for documentation, retrieval, and future assistance. Future work will evaluate the framework and related ego/exo datasets using boundary quality, state-update accuracy, compression ratio, retrieval fidelity, and long-horizon QA. We will also study learned cue fusion, stronger exocentric grounding, privacy-preserving evidence retention, and real-time/on-device variants for practical industrial use.

\bibliographystyle{splncs04}
\bibliography{main}
\end{document}